\RequirePackage[svgnames]{xcolor}

\documentclass[11pt,letterpaper]{mystyle}

\usepackage[all]{hypcap}
\usepackage[svgnames]{xcolor}
\usepackage[comma,authoryear,compress]{natbib}
\usepackage{hyperref}[citecolor=lightblue]

\hypersetup{
    colorlinks = true,
    citecolor = {teal},
}

\usepackage{algorithm}
\usepackage{algorithmicx}
\usepackage{algpseudocode}
\usepackage{microtype}
\usepackage{graphicx}
\expandafter\def\csname ver@subfig.sty\endcsname{}
\usepackage{booktabs} %
\usepackage{float}
\usepackage{bigstrut}

\usepackage{amsmath}
\usepackage{amssymb}
\usepackage{mathtools}
\usepackage{amsthm}
\usepackage{mathrsfs}
\usepackage{nicefrac}
\usepackage{dsfont}
\usepackage{enumitem}
\usepackage{subcaption}
\usepackage{graphicx,subfig}
\usepackage{cleveref}
\usepackage{bxcoloremoji}
\usepackage{datetime2}
\usepackage{float}

\usepackage[utf8]{inputenc} %
\usepackage[T1]{fontenc}    %
\usepackage{hyperref}       %
\usepackage{url}            %
\usepackage{booktabs}       %
\usepackage{amsfonts}       %
\usepackage{nicefrac}       %
\usepackage{microtype}      %
\usepackage{graphicx}
\usepackage{subcaption} 
\usepackage{fdsymbol}
\usepackage{wrapfig}
\usepackage{lipsum}
\usepackage{enumitem}
\usepackage{stackengine}
\usepackage[font=small,labelfont=bf]{caption}
\usepackage{color}
\usepackage{adjustbox}
\usepackage{multirow, multicol}

\usepackage{rotating}
\usepackage{makecell}

\usepackage{tcolorbox}
\tcbuselibrary{listings,skins,breakable}
\usepackage{listings}
\usepackage{subcaption}

\newcommand{\x}{\mathbf{x}}

\definecolor{blanchedalmond}{rgb}{1.0, 0.92, 0.8}
\definecolor{carmine}{rgb}{0.59, 0.0, 0.09}
\definecolor{lightblue}{rgb}{0.22,0.45,0.70}%

\renewcommand{\mathbf}{\boldsymbol}

\makeatletter
\def\Ddots{\mathinner{\mkern1mu\raise\p@
\vbox{\kern7\p@\hbox{.}}\mkern2mu
\raise4\p@\hbox{.}\mkern2mu\raise7\p@\hbox{.}\mkern1mu}}
\makeatother

\definecolor{amaranth}{rgb}{0.9, 0.17, 0.31}
\definecolor{antiquebrass}{rgb}{0.8, 0.58, 0.46}
\definecolor{antiquefuchsia}{rgb}{0.57, 0.36, 0.51}
\definecolor{chromeyellow}{rgb}{0.31, 0.47, 0.26}

\newtcolorbox{AIbox}[2][]{aibox,title=#2,#1}
\definecolor{lightgreen}{rgb}{0.22,0.70,0.30}%
\definecolor{Gray}{gray}{0.95}
\definecolor{Cornsilk}{rgb}{1.0, 0.97, 0.86}

\definecolor{lightblue}{HTML}{0064E0}
\definecolor{fg}{HTML}{1C2B33}
\definecolor{bg}{HTML}{F1F4F7}

\newcommand{\name}{\textsc{Rubric-RM}}
\newcommand{\modelname}{\textsc{Rubric-RM}}

\newcommand{\dataset}{{OpenRubrics}}

\usepackage{amsmath}

\usepackage[all]{hypcap}
\title{\LARGE OpenRubrics: Towards Scalable Synthetic Rubric Generation for Reward Modeling and LLM Alignment}

\runningtitle{Towards Scalable Synthetic Rubric Generation for Reward Modeling and LLM Alignment}

\author{
Tianci Liu\textsuperscript{1,*} \quad
Ran Xu\textsuperscript{2,*} \quad
Tony Yu\textsuperscript{3} \quad
Ilgee Hong\textsuperscript{3} \\
\bf Carl Yang\textsuperscript{2} \quad
Tuo Zhao\textsuperscript{3} \quad
Haoyu Wang\textsuperscript{4} \\
\textsuperscript{1}Purdue University \quad
\textsuperscript{2}Emory University \quad
\textsuperscript{3}Georgia Institute of Technology \quad
\textsuperscript{4}University at Albany
}

\begin{document}

\begin{abstract}
Reward modeling lies at the core of reinforcement learning from human feedback (RLHF), yet most existing reward models rely on scalar or pairwise judgments that fail to capture the multifaceted nature of human preferences.
Recent studies have explored \emph{rubrics-as-rewards} (RaR) that uses structured criteria to capture multiple dimensions of response quality. However, producing rubrics that are both reliable and scalable remains a key challenge.
In this work, we introduce \textbf{\dataset{}}, a diverse, large-scale collection of (prompt, rubric) pairs for training rubric-generation and rubric-based reward models. 
To elicit discriminative and comprehensive evaluation signals, we introduce \emph{Contrastive Rubric Generation} (CRG), which derives both hard rules (explicit constraints) and principles (implicit qualities) by contrasting preferred and rejected responses. We further remove noisy rubrics via preserving preference–label consistency. Across multiple reward-modeling benchmarks, our rubric-based reward model, \modelname{}, surpasses strong size-matched baselines by 8.4\%. These gains transfer to policy models on instruction-following and biomedical benchmarks. 
\vspace{2mm}


\vspace{5mm}

\coloremojicode{1F4C5} \textbf{Date}: \today



\coloremojicode{1F917} \textbf{Model Weights \& Checkpoints}: \href{https://huggingface.co/OpenRubrics/models}{https://huggingface.co/OpenRubrics/models}

\coloremojicode{1F4DA} \textbf{Datasets}: \href{http://huggingface.co/OpenRubrics/datasets}{http://huggingface.co/OpenRubrics/datasets}

\coloremojicode{1F4E7} \textbf{Contact}: \href{mailto:liu3351@purdue.edu}{liu3351@purdue.edu}; \href{mailto:ran.xu@emory.edu}{ran.xu@emory.edu}; \href{mailto:hwang28@albany.edu}{hwang28@albany.edu}

\end{abstract}

\maketitle
\def\thefootnote{$^{*}$}\footnotetext{These authors contributed equally to this work, order was determined randomly (by rolling a die).}\def\thefootnote{\arabic{footnote}}
\vspace{3mm}
\vspace{-2mm}
\section{Introduction}
\label{sec:intro}
\vspace{-1mm}
Reward modeling is central to reinforcement learning from human feedback (RLHF) and is widely used to align large language models (LLMs) with human preferences~\citep{ouyang2022training,wu2023fine,bhaskar2025language}. 
By assigning a scalar score~\citep{ouyang2022training} or preference label~\citep{chen2025rm} to each response, reward modeling provides the optimization signal during training and steers the policy LLM toward generating helpful and harmless responses.

While lots of efforts have been paid on RL with \emph{verifiable reward} (RLVR)~\citep{guo2025deepseek,yue2025does}, many high-value applications of LLMs, such as long-form question answering, general helpfulness, operate in inherently subjective domains where correctness cannot be sufficiently captured by binary signals. 
To bridge this gap, \emph{rubrics-as-rewards (RaR)}~\citep{gunjal2025rubrics} have emerged as a new paradigm for reward modeling. 
Rubrics include structured natural language criteria that decompose quality into interpretable and measurable dimensions, providing a more consistent and transparent evaluation framework than scalar judgments. For policy models, rubrics also enable optimization to be guided by explicit principles.

Despite their great promise, the construction of high-quality rubrics remains an open challenge. Existing benchmarks~\citep{arora2025healthbench} curate rubrics with the effort from domain experts, which is costly and difficult to scale. Recent works~\citep{huang2025reinforcement,viswanathan2025checklists,gunjal2025rubrics} typically generate rubrics via direct prompting LLMs, but those approaches suffer from limited quality control over rubrics and can be prohibitively expensive when relying on commercial APIs.

In this work, we present \dataset{}, a large collection of (prompt, rubrics) pairs to facilitate rubric-generation model training. 
Specifically, we prompt the LLM to generate two complementary types of rubrics: \emph{hard rules}, which capture explicit and objective constraints specified in the prompt, and \emph{principles}, which summarize implicit and generalizable qualities of strong responses. This design allows the rubrics to capture both surface-level requirements and deeper dimensions of quality.
Although hard rules are typically straightforward to extract, the principles are more subtle and require fine-grained reasoning. To address this, we propose \emph{Contrastive Rubric Generation} (CRG), which conditions on user queries paired with both chosen and rejected responses. By leveraging negative contrasts, CRG encourages the model to identify discriminative qualities that distinguish stronger answers from weaker ones, yielding more comprehensive and ranking-aware rubric signals. 
To further ensure reliability and reduce the noise, we apply preference-label consistency through rejection sampling, retaining only rubrics that yield correct preference predictions.

Our contributions are three-fold:
\begin{itemize}[leftmargin=0.4cm]
\item We introduce \dataset{}, a large-scale and diverse collection of rubrics. This dataset enables both rubric generation models and rubric-informed reward modeling at scale.

\item We distinguish between two fundamental types of rubrics and propose a \emph{contrastive rubric generation} strategy that trains models to produce comprehensive and discriminative rubrics from prompts and responses. Besides, we introduce \emph{preference–label consistency} that improves the quality and reliability of the rubric.

\item We conduct extensive experiments on eight benchmark datasets, where \modelname{} consistently outperforms strong baselines by 8.4\%. Moreover, when integrated into policy optimization, \modelname{} consistently yields notable gains on challenging instruction following and medical benchmark.
Case studies further verify the benefits of combining hard rules and principles, showing that rubrics help reduce false positives from overly long outputs.
\end{itemize}
\section{Related Works}
\label{sec:related_work}

\paragraph{Reward Modeling.} 
Standard reward models assign scalar scores to responses by applying a ranking loss under the Bradley–Terry framework~\citep{bradley1952rank,liu2025skywork}. To enhance reasoning capability, generative reward models (GenRMs) incorporate synthesized Chains of Thought (CoT), enabling more accurate reward estimation~\citep{ankner2024critique,yu-etal-2025-self,zhang2025generative,liang-etal-2025-generative}. Beyond the pointwise setting, pairwise reward models have been proposed to directly compare multiple responses~\citep{xu2025unified,liu2025pairjudge}. 
More recently, reinforcement learning has been leveraged to further optimize reward models, enabling them to reason explicitly over comparisons and thereby achieve stronger alignment performance~\citep{chen2025judgelrm,chen2025rm,whitehouse2025j1,guo2025reward}. 
Orthogonal to these efforts, our work focuses on improving reward modeling quality with structured rubrics. By introducing rubric-based evaluation signals, we complement existing approaches with an additional layer of interpretability that yield performance gains.

\paragraph{Rubrics as Rewards.}  
Recent work has explored rubrics for both evaluation and alignment.
Rubrics provide structured assessments of model generations~\citep{arora2025healthbench,hashemi2024llm,pathak2025rubric,lin2025wildbench}, guide instruction following and domain adaptation~\citep{viswanathan2025checklists,gunjal2025rubrics}, improve safety via rule-based rewards~\citep{mu2024rule}, and have been combined with verifiable rewards for reasoning tasks~\citep{huang2025reinforcement,zhou2025breaking}.
Yet most existing approaches rely on prompting frontier LLMs to generate rubrics, which limits scalability and consistency. 
{Our work introduces a more scalable framework for \emph{high quality synthetic rubric generation}}, improving both reward quality and interpretability at a cheaper cost.
Concurrently, \citet{zhang2025chasing} also investigate rubric generation, but focus on \emph{iterative refinement} to mitigate reward over optimization, whereas we emphasize scalable synthesis and rubric–preference consistency.

\section{Preliminaries}
\begin{figure*}[t!]
    \centering
    \includegraphics[width=0.95\textwidth]{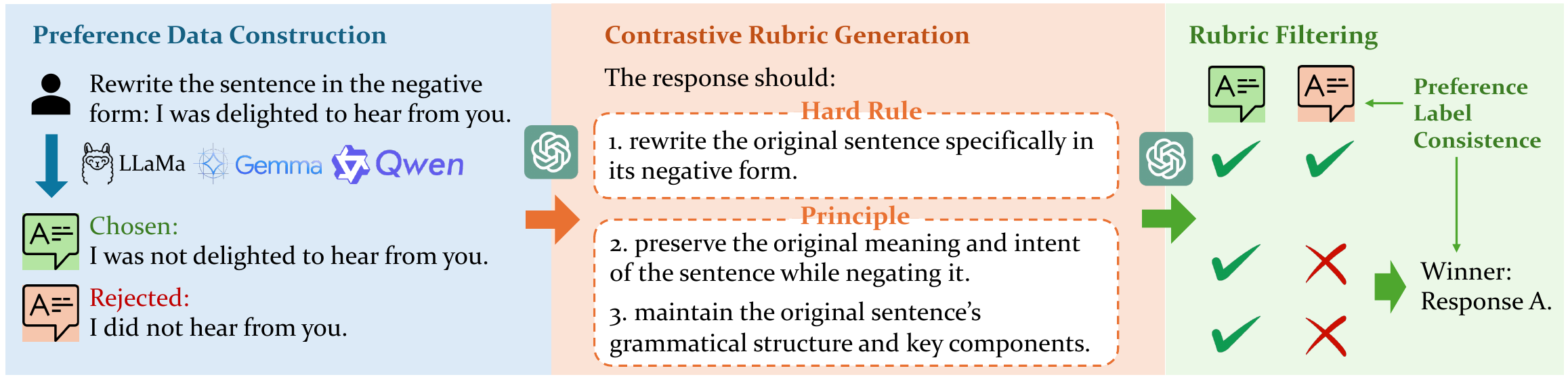}
    \caption{\centering Overall Framework for Synthetic Rubric Generation in \dataset{}. \vspace{-1ex}}
    \label{fig:framework}
\end{figure*}

\noindent \textbf{Rubrics.}
We define rubrics as a structured set of evaluation criteria tailored to a given prompt. 
Formally, let $x$ denote an input prompt and $\hat{y}$ a model-generated response. 
A rubric $\mathcal{R}(x)$ is represented as a collection of $k$ criteria 
$\mathcal{R}(x) = \{c_i\}_{i=1}^k$,
where each $c_i$ denotes a rubric description specifying an aspect of response quality (e.g., factual correctness, reasoning soundness, style).

\noindent \textbf{Rubrics-based Reward Models.}
Building on prior work in pairwise reward modeling~\citep{liu2025pairjudge,chen2025rm,guo2025reward,xu2025unified},  
we focus on a comparative setting where the goal is to evaluate the relative quality of two candidate responses.  
Given a prompt $x$ and two samples $(\hat{y}_1, \hat{y}_2)$,  
a pairwise rubric-based reward function is defined as
\begin{equation}
\setlength{\abovedisplayskip}{5pt}
\setlength{\belowdisplayskip}{5pt}
\nonumber
   \text{reward}_{\text{pair}}(x,\hat{y}_1,\hat{y}_2) 
   = r_{\theta}\big(x,\hat{y}_1,\hat{y}_2;\{c_i\}_{i=1}^k \big),
\end{equation}
where $\text{reward}$ is the binary preference label, $r_\theta$ is the reward model that integrates rubric criteria $\{c_i\}$ when producing a preference judgment.  

The overall framework for \dataset{} is in Figure \ref{fig:framework}. Our overall objective is two-fold:  
(1) constructing a rubric dataset $\mathcal{D}_{\text{rubric}}$ to train a generation model $g_{\theta}$ that automatically synthesizes rubrics $\mathcal{R}(x)$ given a prompt $x$; and  
(2) building a reward modeling dataset $\mathcal{D}_{\text{rm}}$ to train a rubric-guided reward model $r_{\phi}$ capable of producing reliable and interpretable pairwise judgments.  
This two-stage formulation decomposes evaluation into \emph{rubric generation} and \emph{rubric-conditioned reward prediction}, bridging human-aligned criteria and automated preference modeling.

\section{\dataset{}}
\label{sec:dataset}
\subsection{Data Construction}
\label{sec:data_construct}
\textbf{Data Sources.} To generate high-quality rubrics that generalize across tasks and domains, we integrate a range of publicly available preference and instruction-tuning datasets, 
balancing general-domain data with domain-specific resources.  
Specifically, our dataset draws from the following sources:
\begin{itemize}[leftmargin=0.4cm,itemsep=0.1pt]
    \item \textbf{UltraFeedback}~\citep{cui2024ultrafeedback}, which aggregates preference annotations from 
    \emph{Evol-Instruct}~\citep{xu2024wizardlm}, \emph{UltraChat}~\citep{ultrachat}, \emph{ShareGPT}~\citep{sharegpt}, and \emph{TruthfulQA}~\citep{lin-etal-2022-truthfulqa}.
    \item \textbf{Magpie}~\citep{xu2025magpie}: a large-scale synthetic alignment dataset generated by self-prompting LLMs across diverse domains.
    \item \textbf{Skywork-Preference}~\citep{liu2024skywork}, which integrates data from \emph{HelpSteer2}~\citep{wang2024helpsteer} and \emph{OffsetBias}~\citep{offsetbias}.
    \item \textbf{Synthetic-IF}~\citep{lambert2025tulu}, a collection of human preference judgments tailored for verifiable instruction-following.
    \item \textbf{MegaScience}~\citep{fan2025megascience}, a specialized corpus spanning multiple scientific domains including physics, biology, and chemistry.
    \item \textbf{Medical-o1}~\citep{chen2024huatuogpt}, a medical SFT dataset curated for diagnostic reasoning tasks.
\end{itemize}

\begin{figure*}[t!]
    \centering
    \begin{subfigure}{0.47\linewidth}
        \includegraphics[width=\linewidth]{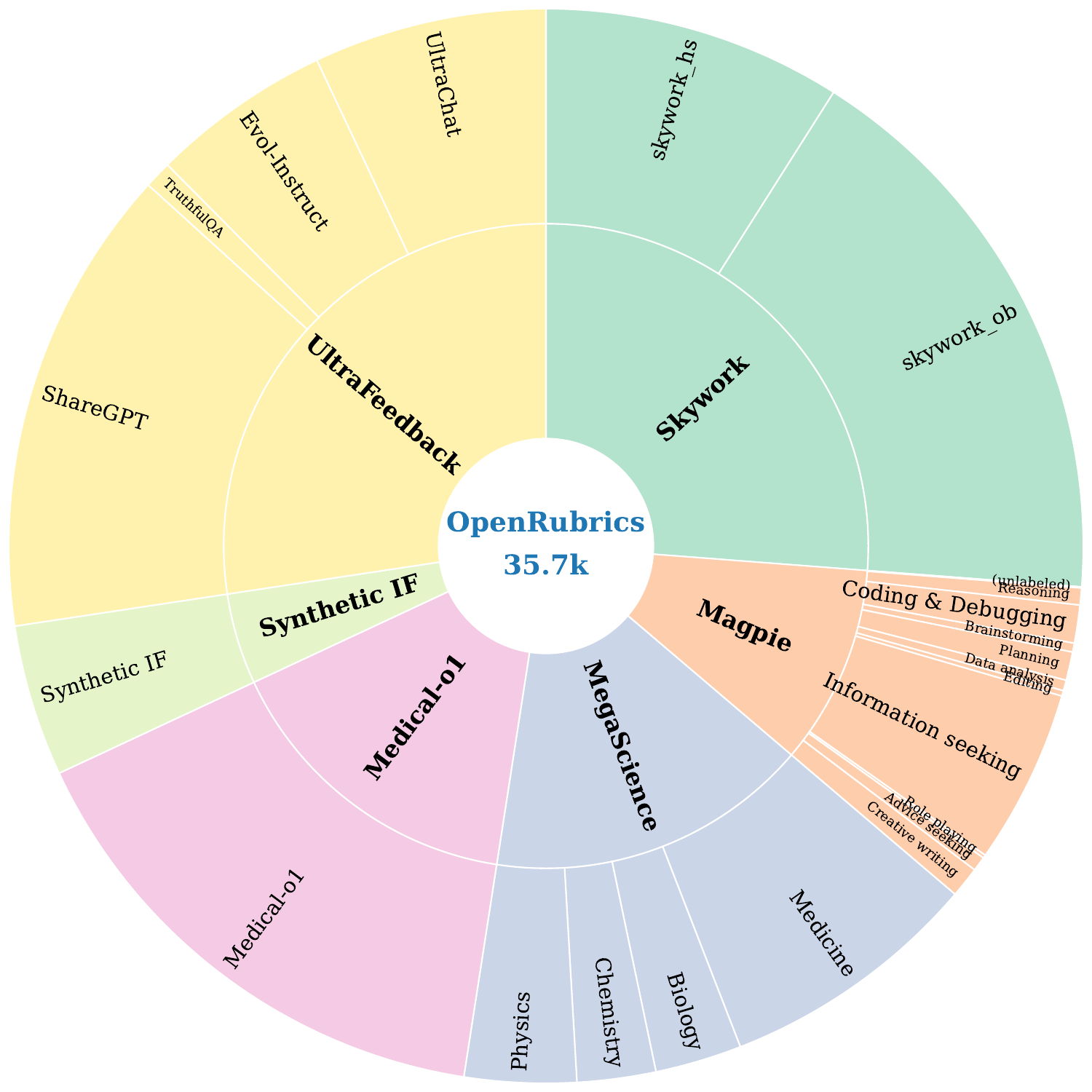}
        \caption{\centering The data distribution for \dataset{} (in \# instructions).}
        \label{fig:pie_chart}
    \end{subfigure}
    \hfill
    \begin{subfigure}{0.51\linewidth}
        \includegraphics[width=\linewidth]{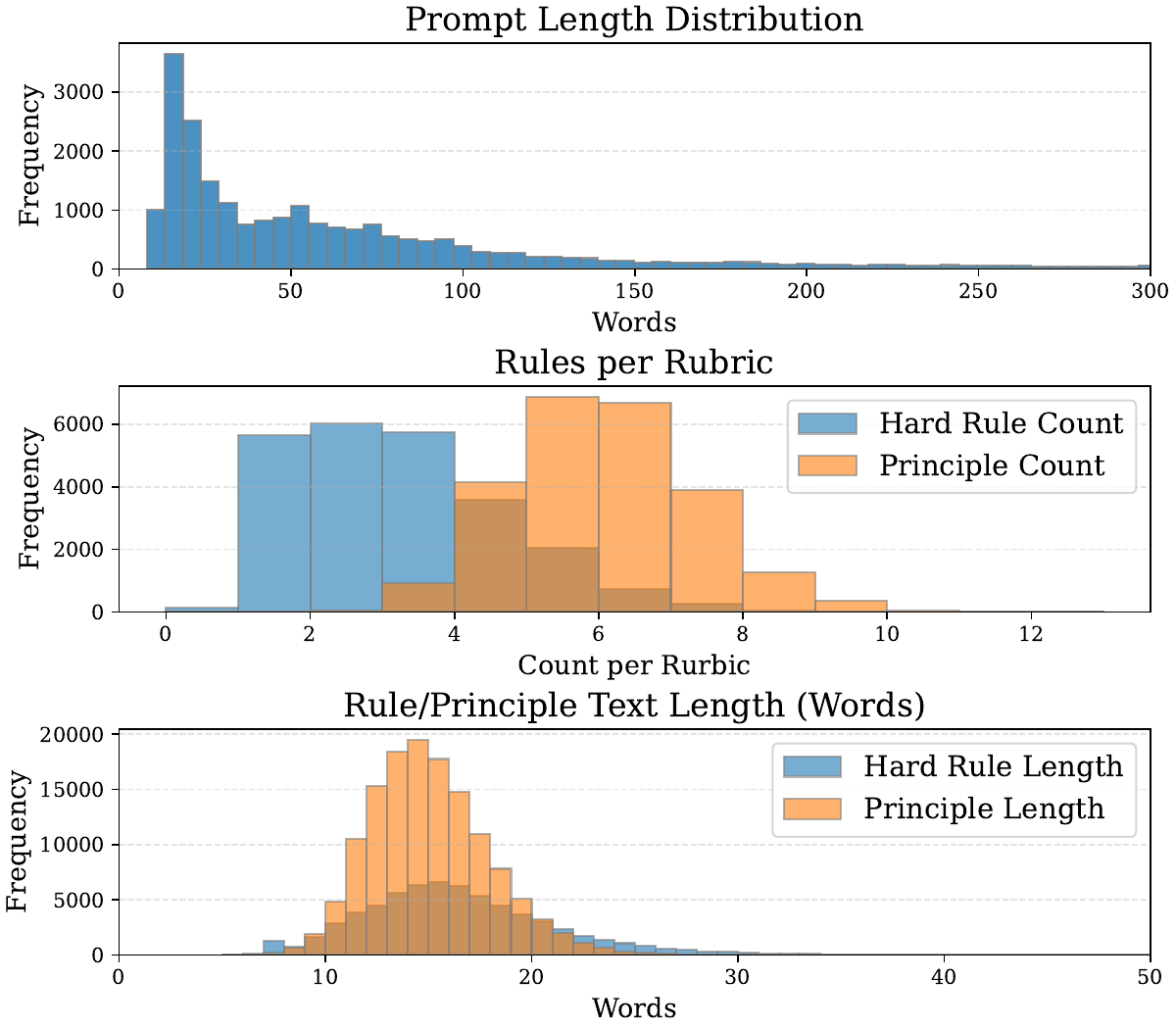}
        \caption{\centering The distribution for the length of prompts and rubrics, as well as the number of rubrics.}
        \label{fig:hist}
    \end{subfigure}
    \caption{\centering Statistics Overview of \dataset{}.\vspace{-2ex}}
    \label{fig:stats}
\end{figure*}

\noindent \textbf{Preference Data Construction.}  
To build preference data for rubric generation and judge training (see Sec.~\ref{sec:rm_train}), we reuse existing preference and SFT datasets with tailored processing.
For \textbf{UltraFeedback}, we select the highest-scoring response as the \emph{chosen} and the lowest as the \emph{rejected}.
For \textbf{MegaScience}, and \textbf{Medical-o1}, we generate multiple responses using \textit{Qwen-3-8B/14B}~\citep{yang2025qwen3}, \textit{Llama-3.1-8B}~\citep{grattafiori2024llama}, and \textit{Gemma-3-12B}~\citep{team2025gemma}, selecting one from each model.
For \textbf{Synthetic-IF}, responses satisfying all verification functions are labeled as \emph{chosen}, and others as \emph{rejected}.
For the \textbf{MegaScience} and \textbf{Medical-o1} datasets, we employ an ensemble of open-source reward models: \textit{Athene-RM-8B}~\citep{Athene2024} and \textit{Skywork-Reward-V2-Llama-3.1-8B-40M}~\citep{liu2025skywork} to rank responses and form best–worst preference pairs.

    



\subsection{Rubrics Synthesis}
\label{sec:rubrics_gen}
After collecting a diverse set of preference pairs, our objective is to construct a set of rubrics that serve as \emph{anchors} to guide   reward modeling.
To comprehensively represent different types of constraints while preserving discriminative granularity, we categorize rubrics into two  types:
(1) \textbf{Hard Rules}, which capture explicit requirements stated in the user’s prompt; and
(2) \textbf{Principles}, which describe higher-level qualitative aspects such as reasoning soundness, factuality, or stylistic coherence.
We then introduce two strategies for generating high-quality rubrics as follows:  


\noindent \textbf{Contrastive Rubric Generation.} 
Given a preference dataset
$
\mathcal{D}=\left\{\left(x_i, \{\hat{y}_{i,m}\}_{m=1}^{M_i}, \{\ell_{i,m}\}_{m=1}^{M_i}\right)\right\}_{i=1}^N,
$
where $x_i$ is the prompt, $\{\hat{y}_{i,m}\}_{m=1}^{M_i}$ and $\{\ell_{i,m}\}_{m=1}^{M_i}$ denotes a list of candidate responses and their corresponding preference signals, respectively.
The preference signal $\ell_{i,m}$ can be a real-valued score when available, or a binary chosen--rejected label.
We unify both cases by constructing a strictly ordered list in descending preference:
$
\hat{y}_{i,1}\succ \hat{y}_{i,2}\succ \cdots \succ \hat{y}_{i,M_i},
$
where the order is induced by $\ell$.
In particular, when only chosen--rejected labels are available, we simply set $M_i=2$ with $\hat{y}_{i,1}$ as chosen and $\hat{y}_{i,2}$ as rejected.
Our objective is to infer rubrics $\mathcal{R}(x_i)$ that capture qualities a good response should satisfy and the criteria that explain preference differences across the list, leveraging rich and fine-grain supervision from the comparisons.
Formally, we prompt an instruction-tuned LLM $h_\psi$ as
$
\mathcal{R}(x_i) \sim h_\psi\!\left(x_i, \{\hat{y}_{i,m}\}_{m=1}^{M_i}, \{\ell_{i,m}\}_{m=1}^{M_i}\right).
$
The generator is asked to produce a set of discriminative evaluation criteria
$\mathcal{R}(x_i)=\left\{c_{i,1}, \ldots, c_{i,k_i}\right\}$,
where each $c_{i,j}$ describes a specific aspect and is expected to distinguish higher-preference responses from lower-preference ones.
This listwise setting encourages the model to discover rubric dimensions that are both task-sensitive and preference-aligned, while exploiting any available fine-grained preference information.

\noindent \textbf{Rubric Filtering with Preference-label Consistency.}
Not all generated rubrics faithfully capture the human preference signal.
To ensure reliability, we perform a consistency-based filtering step by prompting the same instruction-tuned LLM $h_\psi$ again.
Specifically, for each prompt $x_i$ with an ordered response list $\{\hat{y}_{i,m}\}_{m=1}^{M_i}$, we construct a set of induced pairwise comparisons
$
\mathcal{P}_i=\{(a,b)\mid 1\le a<b\le M_i\},
$
where the human preference label for each pair is $\ell_{i,(a,b)}=1$ (i.e., $\hat{y}_{i,a}\succ \hat{y}_{i,b}$).
When the original data only provides chosen--rejected labels, we have $M_i=2$ and thus $|\mathcal{P}_i|=1$.
Next, we concatenate the full rubric $\mathcal{R}(x_i)$ into the context and ask the model to judge each induced pair:
\begin{equation}
\hat{l}_{i,(a,b)} = h_\psi\!\left(x_i, \mathcal{R}(x_i), \hat{y}_{i,a}, \hat{y}_{i,b}\right),
\end{equation}
where $\hat{l}_{i,(a,b)}=(\hat{r}_{i,(a,b)}, \hat{\ell}_{i,(a,b)})$ denotes the prediction rationale and the predicted preference, respectively.
A rubric is considered reliable for prompt $x_i$ only if it yields preference predictions consistent with human labels on the induced pair set:
\begin{equation}
\mathrm{Acc}_i
=
\frac{1}{|\mathcal{P}_i|}
\sum_{(a,b)\in\mathcal{P}_i}
\mathbb{I}\!\left[\hat{\ell}_{i,(a,b)}=\ell_{i,(a,b)}\right]
\ \ge\ \tau,
\end{equation}
where $\tau$ is a threshold (we use $\tau=0.5$).
Finally, for each induced pair $(a,b)\in\mathcal{P}_i$, we retain the rubric-conditioned training instance only if
(i) the rubric passes the group-level verification ($\mathrm{Acc}_i\ge\tau$), and
(ii) the current pairwise prediction is correct ($\hat{\ell}_{i,(a,b)}=\ell_{i,(a,b)}$).
Formally,
\begin{equation}
\mathcal{R}^*_{i,(a,b)} =
\begin{cases}
\mathcal{R}(x_i), & \text{if} \quad \mathrm{Acc}_i \ge \tau \quad \text{and} \quad \hat{\ell}_{i,(a,b)}=\ell_{i,(a,b)},\\
\emptyset, & \text{otherwise.}
\end{cases}
\end{equation}
This yields a collection of high-quality rubrics that are both interpretable and empirically consistent with human preferences.
The final \emph{rubrics-conditioned preference dataset} is constructed as a pairwise one
\begin{align}
\mathcal{D}_{\text{rubric}}
&=
\left\{
\left(x_i, \hat{y}_{i,a}, \hat{y}_{i,b}, \ell_{i,(a,b)}, \mathcal{R}^*_{i,(a,b)}\right)
\ \middle|\ 
i\in[N],\ (a,b)\in\mathcal{P}_i,\ \mathcal{R}^*_{i,(a,b)}\neq\emptyset
\right\} \\
&\triangleq
\{(x_i,\hat{y}_i^+,\hat{y}_i^-,\mathcal{R}^*(x_i))\}_{i=1}^M.
\end{align}
Namely, we flatten and re-index all retained pairs to obtain the pairwise dataset,
where new instance corresponds to one retained $(i,(a,b))$ with $\hat{y}_i^+=\hat{y}_{i,a}$ and $\hat{y}_i^-=\hat{y}_{i,b}$.

\begin{figure}[t]
  \centering
   \vspace{-1ex}
  \includegraphics[width=0.7\linewidth]{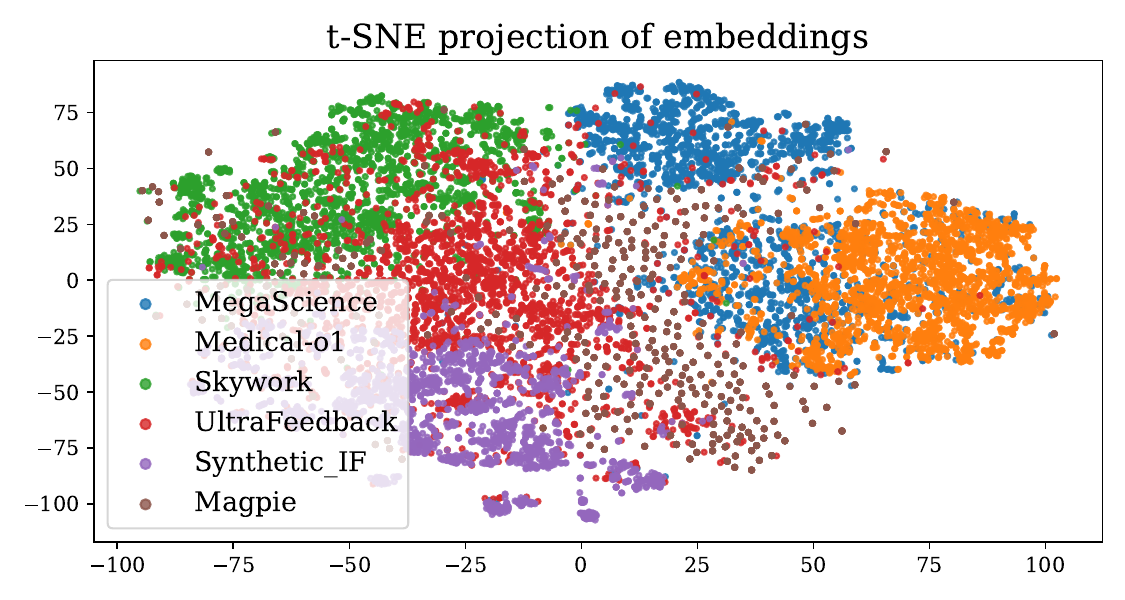}
  \caption{The T-SNE plot for embeddings of prompts. \vspace{-2ex}}
  \label{fig:tsne_plot}
\end{figure}

\noindent \textbf{Rubric Statistics Overview.}
We analyze the curated rubric set along three axes: (1) domain coverage (instruction following, reasoning, general helpfulness; Figure~\ref{fig:pie_chart}); (2) the balance between \emph{hard rules} and \emph{principles} as well as the length of prompts and rubrics (Figure~\ref{fig:hist}); and (3) semantic diversity of prompt topics, visualized via t-SNE on embeddings from Qwen-3-Embedding-0.6B~\citep{zhang2025qwen3} (Figure~\ref{fig:tsne_plot}). 
These statistics confirm that the synthesized rubrics provide comprehensive, yet discriminative coverage, forming a foundation for rubric-based reward modeling.

\subsection{Reward Model Training and Inference}
\label{sec:rm_train}
After collecting the rubrics-based dataset, we proceed to develop a rubric generation model that outputs evaluation rubrics and a reward model \modelname{} that generates final preference labels.

\noindent \textbf{Rubric Generation.}
We first fine-tune $g_\theta$ to generate rubrics $\mathcal{R}^*$ conditioned on the prompt $x$. Given the dataset $\mathcal{D}_{\text{rubric}} = \{(x_i, \hat{y}_i^{+}, \hat{y}_i^{-}, \mathcal{R}^*(x_i))\}_{i=1}^{M}$, we optimize the standard cross-entropy objective:

\begin{small}
\begin{equation}
\setlength{\abovedisplayskip}{-5pt}
\setlength{\belowdisplayskip}{5pt}
\mathcal{L}_{\text{SFT}}^{\text{rubric}}
= -\,\mathbb{E}_{(x, \hat{y}^{+}, \hat{y}^{-}, \mathcal{R}^*) \in \mathcal{D}_{\text{rubric}}}
\sum_{t=1}^{|\mathcal{R}^*|}
\log p_\theta \big(\mathcal{R}^*_t \mid x, \mathcal{R}^*_{<t}\big).
\nonumber
\end{equation}
\end{small}

\noindent \textbf{Reward Model Training.}
We then train the reward model $r_\phi$ to predict preference labels $\hat{l}$ using the generated rubrics. Using $\mathcal{D}_{\text{rm}} = \{(x_i, \hat{y}_i^{+}, \hat{y}_i^{-}, \mathcal{R}^*(x_i), \hat{l}_i)\}_{i=1}^{M}$, we optimize $r_\phi$ to predict $\hat{l}$ conditioned on the prompt, response pair, and rubric:

\begin{small}
\begin{equation}
\setlength{\abovedisplayskip}{-8pt}
\setlength{\belowdisplayskip}{5pt}
\mathcal{L}_{\text{SFT}}^{\text{rm}}
= -\,\mathbb{E}_{\mathcal{D}_{\text{rm}}}
\sum_{t=1}^{|\hat{l}|}
\log p_\phi \big(\hat{l}_t \mid x, \hat{y}^{+}, \hat{y}^{-}, \mathcal{R}^*(x), \hat{l}_{<t}\big).
\nonumber
\end{equation}
\end{small}


\noindent \textbf{Inference.}
At inference time, given a pairwise test instance $(x, y^{\text{A}}, y^{\text{B}})$,  
\modelname{} performs a two-stage process to predict the final preference label:
(1) the rubric generator produces (or retrieve a cached) rubric conditioned on the instruction $x$ as $\hat{\mathcal{R}}(x) = g_\theta(x)$; (2) the reward model then predicts the verdict over candidate responses $(y^{\text{A}}, y^{\text{B}})$ conditioned on the generated rubric from two possible labels $\mathcal{C} = \{\text{A is better}, \text{B is better}\}$:

%
\begin{small}
\begin{equation}
\setlength{\abovedisplayskip}{4pt}
\setlength{\belowdisplayskip}{5pt}
\hat{l}
= \arg\max_{k \in \mathcal{C}}
p_\phi \big(k \mid x, y^{\text{A}}, y^{\text{B}}, \hat{\mathcal{R}}(x)\big).
\nonumber
\end{equation}
\end{small}
This ensures that the judgment from \modelname{} is explicitly grounded in rubric criteria.



\section{Experiment}







\subsection{Datasets and Experiment Settings}
\label{subsec:datasets-settings}

\noindent \textbf{Training data.}
We train both components of {\name}: the \emph{rubric generator} and the \emph{judge}, on the curated {\dataset} as presented in Sec.~\ref{sec:rubrics_gen}. Rubrics are produced with contrastive signals from chosen/rejected responses and filtered by preference–label consistency before use. Unless otherwise noted, we use the science-related slice of OpenRubrics to better match our domain study on HealthBench/medical evaluation.

\noindent \textbf{Backbone and variants.}
Both the rubric generator and the judge are fine-tuned from Qwen-3-8B (``{\name}-8B'') unless specified. At inference time, {\name} follows the two-stage process as detailed in Sec \ref{sec:rm_train}. We also report an ensemble variant, \textit{voting@5}, which aggregates five independently sampled judge trajectories by majority vote.

\noindent \textbf{Baselines.}
We compare {\name} against strong, same-scale white-box reward (judge) models: JudgeLRM-7B~\citep{chen2025judgelrm}, RRM-7B~\citep{guo2025reward}, and RM-R1-7B~\citep{chen2025rm}.
We also report larger RM-R1-14B~\citep{chen2025rm} and API judges for reference when available. To isolate the benefit of rubric-aware fine-tuning, we include naive {Qwen-3-8B (Rubric+Judge)} that directly prompts the base model to produce rubrics and then make judgments.

\noindent \textbf{Evaluation benchmarks and metrics.}
We evaluate {\name} as a pairwise reward model on popular reward-modeling benchmarks: \emph{RewardBench} (Chat, Chat-Hard)~\citep{rewardbench}, \emph{RM-Bench}~\citep{liu2025rmbench}, \emph{PPE-IFEval}~\citep{ppe}, \emph{FollowBench}~\citep{followbench}, \emph{InfoBench}~\citep{infobench}, \emph{IFBench}~\citep{ifbench}, and \emph{RewardBench2} (Precise-IF, Focus)~\citep{malik2025rewardbench2}. 
While \emph{FollowBench} and \emph{InfoBench} were originally designed to assess instruction-following capabilities of LLMs, we adapt them to pairwise evaluation settings by sampling two responses from the same model (Qwen-3-8B/14B), where one response adheres to all specified constraints and the other violates some of them. 
For the domain study we additionally report HealthBench/medical results. We follow each benchmark’s official splits and scoring rules, reporting accuracy, win-rate or other specific scores.

Due to space limits, additioanl implementation details are deferred to Appendix \ref{apd:implementation_details}.

\begin{table*}[!t]
\centering
\renewcommand\arraystretch{0.95}
\resizebox{0.98\linewidth}{!}{%
\begin{tabular}{l cc cccc c cc cc}
\toprule
\multirow{2.5}{*}{} 
& \multicolumn{2}{c}{\bf RewardBench} 
& \multicolumn{4}{c}{\bf IF Evaluation Benchmarks} 
& \bf RM-Bench
& \multicolumn{2}{c}{\bf RewardBench2}
& \multirow{2.5}{*}{\bf HelpSteer3}
& \multirow{2.5}{*}{\bf Avg.} \\ 
\cmidrule(lr){2-3}  \cmidrule(lr){4-7} \cmidrule(lr){8-8} \cmidrule(lr){9-10}
                            & Chat & Chat Hard  & FollowBench   &   PPE-IFEval  &   InfoBench   &   IFBench     &    Chat        & Precise IF    &   Focus  &      &   \\ 
\midrule
\multicolumn{11}{l}{\it Black-box LLMs (For reference only)} \\
\midrule
Claude-3.5-Sonnet           & 96.4 & 74.0       & --            & 58.0          & --            & --            & 62.5           & 38.8          & 87.0     & --   & - \\
API (Rubric+Judge)          & 79.6 & 79.2       & 83.2          & 61.0          & 82.2          & 66.2          & 67.9           & 42.5          & 79.6	    & 71.4 & 71.3 \\
API (direct Judge)          & 89.6 & 71.2       & 81.7          & 59.2          & 72.9          & 60.4          & 67.2           & 13.2          & 63.4     & 70.3 & 64.9 \\ 
\midrule
\multicolumn{11}{l}{\it Larger White-box LLMs (For reference only)} \\
\midrule
RM-R1-14B (Qwen-2.5-Inst)   & 73.5 & 79.8	    & 84.0	        & 59.0          & 85.5	        & 60.8          & 73.2           & 23.8	         & 84.6	    & 74.8 & 69.9 \\
RM-R1-14B (DeepSeek-Dist)   & 90.3 & 78.9       & 89.9          & 61.2          & 82.4          & 59.0          & 71.4           & 30.6          & 79.0     & 74.6 & 71.7 \\
\midrule
\multicolumn{11}{l}{\it White-box Judge/Reward LLMs} \\
\midrule
JudgeLRM-7B                 & \bf 92.1 & 56.1   & 79.8          & 46.0          & 62.7          & 47.5          & 55.4           & 9.4           & 29.1     & 60.2 & 53.8 \\
RRM-7B	                    & 77.7 & 69.5       & 65.5	        & 51.0	        & 68.2	        & 53.2	        & 59.9           & 10.0	         & 60.4	    & 62.4 & 57.8 \\
RM-R1-7B (Qwen-2.5-Inst)    & 83.0 & 70.0	    & 56.3          & 55.2	        & 71.3	        & 55.2          & 64.2           & 20.6	         & 76.2 	& 65.2 & 61.7 \\
RM-R1-7B  (DeepSeek-Dist)   & 85.3 & 67.3       & 69.7          & 51.0          & 70.3          & 56.5          & 62.2           & 13.8          & 55.4     & 62.6 & 59.4 \\
Qwen-3-8B (Rubric+Judge)    & 74.2 & 64.2	    & 71.3          & 57.0          & 74.3	        & 59.5	        & 63.9           & 8.1	         & 44.0     & 60.8 & 57.7 \\

\midrule
\multicolumn{11}{l}{\it \modelname{} (Our proposed model)} \\
\midrule
\rowcolor{lightgreen!10}
{\name-4B}                  & 87.2 & 68.9       & 78.0          & 64.8          & 79.3          & 63.2          & 62.6           & 35.6          & 79.6     & 64.7    & 68.4 \\
\rowcolor{lightgreen!10}
{\name-4B}-voting@5         & 88.7 & 70.2       & 80.7          & 66.2          & 83.0          & 64.9          & 63.8           & 38.1          & 82.6     & 65.0    & 70.3 \\

\rowcolor{lightblue!10}
{\name-8B}                  & 88.2 & 74.1       & 76.1          & 67.0          & 80.8          & 65.4          & 65.7           & 34.4          & 82.2     & 67.0 & 70.1 \\
\rowcolor{lightblue!10}
{\name-8B}-voting@5         & 89.9 &  \bf 75.4   & \bf 81.5          &  \bf 70.8      &  \bf 83.8      &  \bf 67.1      & \bf 67.0           & \bf 40.0      & \bf 86.5 &  \bf 67.5 & \bf 73.0\\
\bottomrule
\end{tabular}
}
\caption{Comparison of different judge and reward models across multiple benchmarks. 
RewardBench2 reports results on Precise IF, and Focus dimensions. 
Rubric API uses GPT-4.1-Mini, and Judge API uses Gemini-2.5-Flash-Lite. 
Best results are highlighted in \textbf{bold}. \vspace{-1ex}
}
\label{tab:main_results}
\end{table*}


\subsection{Performance of {\name}}
We first validate the performance of {\name} for reward modeling. For a more systematic evaluation, we test both 4B and 8B variants of {\name}, which use Qwen3-4B/8B as backbones. 
Table~\ref{tab:main_results} reports the results of {\name}. 

\noindent \textbf{Outperforming Comparable Reward Models.} 
Both {\name}-4B and 8B surpass all existing 7B-scale white-box baselines (e.g., JudgeLRM, RRM, RM-R1). Notably, {\name}-4B (\textbf{68.4}) outperforms the strongest 7B competitors (max 61.7), with {\name}-8B further improving to \textbf{70.1}. This confirms that rubric-aware training yields more reliable signals than generic preference learning, even with reduced parameter counts. \\
\noindent \textbf{Majority Voting Further Enhances Performance.} 
We also evaluate {\name}-voting@5, which aggregates predictions via majority voting across five independent judge trajectories.
{\name}-4B-voting@5 reaches \textbf{70.3}, and {\name}-8B-voting@5 achieves the best overall average of \textbf{73.0}, surpassing much larger models such as RM-R1-14B (71.7) and the Rubric+Judge API (71.3).
These results highlight the stability benefits of rubric-based ensembles. \\
\noindent \textbf{Effectiveness of Rubric-Aware Fine-Tuning.} 
Directly using Qwen-3-8B to generate rubrics and judge yields only 57.7. In contrast, {\name} achieves \textbf{70.1}. This \textbf{+12.4} gain validates our core contribution: high-quality rubrics derived via \textit{contrastive generation} and \textit{consistency filtering} are essential for effective reward modeling. \\
\noindent \textbf{Strength on IF Benchmarks.}
{\name} excels on benchmarks requiring fine-grained instruction adherence. On FollowBench and InfoBench, we achieve \textbf{81.5} and \textbf{83.8} respectively, substantially outperforming baselines like JudgeLRM and RRM. This demonstrates that rubrics capture nuanced constraints better than scalar reward models. \\
In the remaining experiments, we use {\name-8B} as our reward model unless specified.

\begin{table}[t!]
\centering
\renewcommand\arraystretch{0.95}
\resizebox{0.7\linewidth}{!}{%
\arrayrulecolor{black!50}
\begin{tabular}{l|cc|cc|c|c}
\toprule
\multirow{2}{*}{\textbf{Model}} 
& \multicolumn{2}{c|}{\textbf{IFEval (Prompt)}} 
& \multicolumn{2}{c|}{\textbf{IFEval (Inst.)}} 
& {\textbf{IFEval}} 
& {\textbf{InfoBench}} \\ 
\cmidrule(lr){2-3}  \cmidrule(lr){4-5} \cmidrule(lr){6-6} \cmidrule(lr){7-7}
& \textbf{Loose} & \textbf{Strict} & \textbf{Loose} & \textbf{Strict} & \textbf{AVG} &  \textbf{AVG} \\ 
\midrule
GPT-4 (0314)                              & 79.3 & 76.9 & 85.4 & 83.6 & 81.3 & 87.3 \\ 
AutoIF~\citep{dong2024self}               & 56.9 & 47.1 & 67.0 & 57.6 & 57.2 & 80.6 \\
UltraIF~\citep{an2025ultraif}             & 75.4 & 71.3 & 83.0 & 79.4 & 77.3 & 80.7 \\
\midrule
Qwen2.5-7B-Instruct          & 75.0 & 72.5 & 81.8 & 79.9 & 77.3 & 78.1 (\underline{76.0}) \\
+ SFT (Distilled)            & 66.8 & 64.1 & 75.3 & 72.8 & 69.8 & 72.5 \\
+ DPO (via Skywork)          & 75.7 & 68.0 & 83.2 & 78.5 & 76.0 & 82.0 \\
+ DPO (via ArmoRM)           & 73.8 & 70.2 & 81.7 & 78.3 & 76.0 & 83.5 \\
+ DPO (via Ultrafbk.)        & 71.5 & 69.1 & 79.9 & 77.7 & 74.6 & 80.0 \\
+ DPO (via AI Judge)         & 73.0 & 68.9 & 80.9 & 77.8 & 75.2 & 76.1 \\
+ DPO (via RLCF)             & 77.3 & 72.6 & 84.1 & 80.3 & 78.6 & \textbf{84.1} (\underline{81.5}) \\
\midrule
\rowcolor{lightblue!10}
+ DPO (via \name)            & \textbf{78.2} & \textbf{73.9} & \textbf{84.5} & \textbf{81.2} & \textbf{79.5} & 83.0 \\ 
\bottomrule
\end{tabular}%
}
\caption{
Comparison of trained policy models with different reward models on a format-based instruction-following benchmark (IFEval) and an open-ended benchmark (InfoBench). 
Baseline results are from \citet{viswanathan2025checklists}. 
Results with \underline{underlines} are reproduced by us using official checkpoints and evaluation scripts. 
Best scores are in \textbf{bold}.
}
\label{tab:ifeval}
\end{table}

\begin{table}[t]
\centering
\renewcommand\arraystretch{0.95}
\resizebox{0.7\linewidth}{!}{%
\begin{tabular}{l|cc|cc|c}
\toprule
\multirow{2}{*}{\textbf{Model}} 
& \multicolumn{2}{c|}{\textbf{Arena-Hard}} 
& \multicolumn{2}{c|}{\textbf{AlpacaEval}} 
& \multirow{2}{*}{\textbf{AVG}} \\ 
\cmidrule(lr){2-3} \cmidrule(lr){4-5}
& \textbf{Vanilla} & \textbf{SC} 
& \textbf{Vanilla} & \textbf{LC} &  \\ 
\midrule
GPT-4 (0314)                 & 50.0 & 50.0 & 22.1 & 35.3 & 39.4 \\ 
UltraIF~\citep{an2025ultraif}   & 31.4 & -- & -- & -- & -- \\
\midrule
Qwen2.5-7B-Instruct          & 51.3 & 42.8 & 33.5 & 36.2 & 41.0 \\
+ SFT (Distilled)            & 32.6 & 29.2 & 36.1 & 33.3 & 32.8 \\
+ DPO (via Skywork)          & 55.1 & 50.3 & 44.8 & \bf 41.5 & 47.9 \\
+ DPO (via ArmoRM)           & 50.8 & 46.4 & 37.6 & 38.1 & 43.2 \\
+ DPO (via Ultrafbk.)        & 52.8 & 47.9 & 33.7 & 38.7 & 43.3 \\
+ DPO (via AI Judge)         & 51.0 & 44.4 & 28.8 & 33.4 & 39.4 \\
+ DPO (via RLCF)             & \bf 54.6 & 48.4 & 36.2 & 37.1 & 44.1 \\ 
\midrule
\rowcolor{lightblue!10}
+ DPO (via \name)            & 52.9 & \textbf{53.1} & \textbf{47.0} & 41.3 & \textbf{48.6} \\ 
\bottomrule
\end{tabular}%
}
\caption{
Comparison of different alignment strategies applied to Qwen2.5-7B-Instruct on \textbf{Arena-Hard} and \textbf{AlpacaEval}. 
Baseline results are from \citet{viswanathan2025checklists}. 
SC and LC stands for Style-controlled and Length-controlled. 
Best results are in \textbf{bold}.
}
\label{tab:alpaca}
\end{table}

\begin{figure}[t!]
    \centering
    \includegraphics[width=0.99\linewidth]{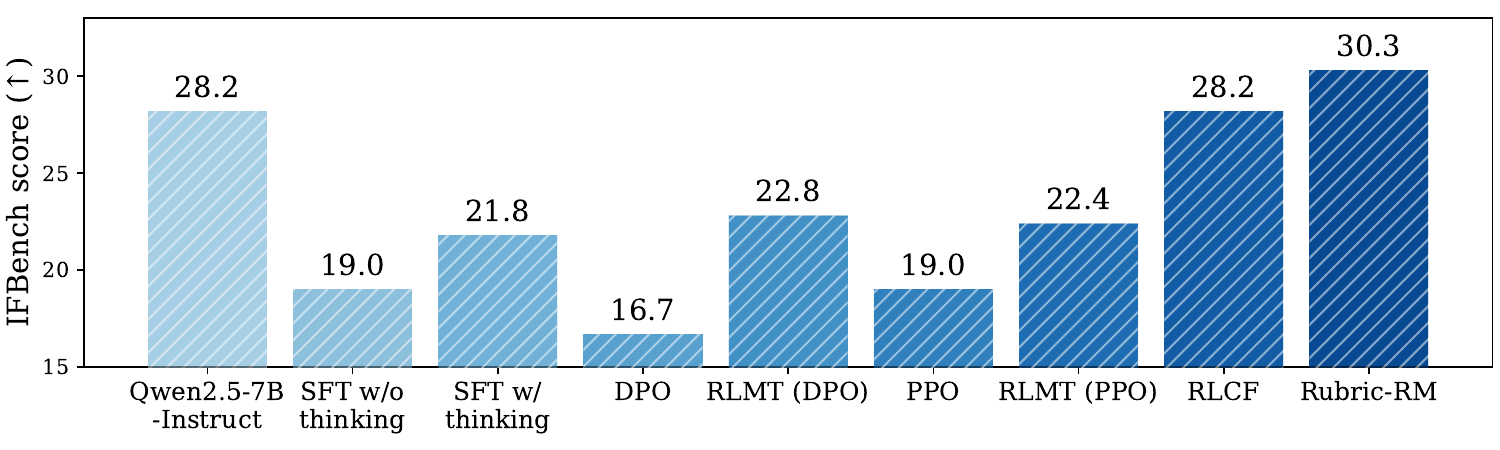}
    \caption{Comparison of policy models on IFBench. Results of baselines except RLCF are from \citet{bhaskar2025language}. We evaluate RLCF with its official checkpoint. \vspace{-1.5ex}}
    \label{fig:ifbench}
\end{figure}

\begin{table}[h!]
\centering
\renewcommand\arraystretch{0.99}
\resizebox{0.7\linewidth}{!}{%
\begin{tabular}{@{}l|cccccc@{}}
\toprule
\textbf{Method} & \textbf{Creative} & \textbf{Planning} & \textbf{Math} & \textbf{Info seeking} & \textbf{Coding} & \textbf{WB Score} \\ \midrule
Claude-3.5-Sonnet$^{*}$                       & 55.6     & 55.6     & 50.2 & 55.5         & 56.5   & 54.7     \\
GPT-4-turbo$^{*}$                             & 58.7     & 56.2     & 51.0 & 57.2         & 55.1   & 55.2     \\
GPT-4o-mini$^{*}$                           & 60.1     & 58.2     & 54.0 & 57.4         & 57.2   & 57.1     \\ \midrule
Qwen2.5-7B-Instruct$^{*}$                     & 50.1     & 51.8     & 47.1 & 50.7         & 45.0   & 48.7     \\
+DRIFT$^{*}$                                  & 52.5     & 53.2     & 50.6 & 52.4         & 50.3   & 51.7    \\
+DPO (via RLCF)                         & 51.4     & 52.7     & 49.0 & 51.3         & 48.8   & 50.5     \\
+DPO (via RLMT (PPO))                    & 52.1        & 52.6        & 45.2    & 51.4            & 48.3      & 49.7       \\ \midrule
\rowcolor{lightblue!10}
+DPO (via {\name})                    & \bf 54.8        &\bf 55.5        &\bf 51.5    &\bf 54.1            &\bf 52.9      &\bf 53.6      \\ \bottomrule
\end{tabular}%
}
\caption{
Comparison of different alignment strategies applied to Qwen2.5-7B-Instruct on \textbf{WildBench}. 
Results are reported for task-specific scores and task macro WB score. 
Baseline results with "$*$" are from \cite{wang2025drift}. We evaluate RLCF and RLMT with their official checkpoints.
Best results are in \textbf{bold}.
}
\label{tab:wildbench}
\end{table}
\subsection{Policy Models with {\name}}
\subsubsection{Instruction-Following Evaluation}
We further evaluate the effectiveness of using {\name} as a reward model for policy optimization on instruction-following tasks, including IFEval, InfoBench, and IFBench. The results are shown in Table~\ref{tab:ifeval} and Figure~\ref{fig:ifbench}.  \\ 
\noindent \textbf{Improved Performance on IFEval and InfoBench.} 
Using {\name} yields the best overall performance among open-source reward-model baselines: the resulting policy model reaches \textbf{79.5} on IFEval (vs.\ 76.0 with Skywork/ArmoRM) and \textbf{83.0} on InfoBench, approaching much larger commercial systems. These results indicate that \emph{rubric-based rewards provide more reliable optimization signals} for constrained instruction following.


\noindent \textbf{Clear Gains on Complex Instruction Following (IFBench).} 
Figure~\ref{fig:ifbench} shows that the policy model optimized with {\name} achieves \textbf{30.3} on IFBench, substantially higher than RLCF (28.2) and RLMT-based methods (22.4--22.8). 
Compared with both supervised fine-tuning variants and reinforcement learning baselines, {\name} provides stronger inductive biases, enabling policies to better capture fine-grained instruction adherence.

Overall, these results demonstrate that {\name}, when used as a reward model, provides a substantially stronger training signal that improves the instruction-following capability of learned policies.




\subsubsection{General Alignment Evaluation}
We evaluate policy model trained with {\name} on alignment benchmarks Arena-Hard, AlpacaEval, and WildBench (Tables~\ref{tab:alpaca} and \ref{tab:wildbench}). 

With DPO optimization, {\name} achieves the best overall average score (\textbf{48.6}) among all open-source reward models. 
On Arena-Hard (style-controlled), it obtains \textbf{53.1}, outperforming Skywork (50.3), Ultrafeedback (47.9), and RLCF (48.4). 
On AlpacaEval, it reaches \textbf{47.0}, surpassing Skywork (44.8) and ArmoRM (37.6). 
On WildBench, aligning Qwen2.5-7B-Instruct with {\name} yields the best macro WB score of \textbf{53.6}, improving over the base model (48.7) by +4.9 and outperforming DPO with RLCF (50.5), RLMT (49.7), and DRN (51.7), with consistent gains across all task categories.
These results show that rubric-based signals provide reliable gains across both vanilla and controlled settings.


\subsection{{\name} for BioMedical Domain}

In this section we further study the effectiveness of {\name} in more specialized biomedical domain, following \citet{arora2025healthbench}.
The Rubric and Judge are \texttt{Qwen-3-8B} backbones fine-tuned with {\dataset} data from science-related domains.

\noindent \textbf{Performance on HealthBench.}
{\name} outperforms comparable-size reasoning reward models on HealthBench, achieving \textbf{68.3}, exceeding RRM-7B (63.3) and RM-R1-7B variants (55.4/66.9), and approaching RM-R1-14B (69.9).
Majority voting further boosts performance: {\name}-{voting@5} reaches \textbf{72.9}, narrowing the gap to 14B reasoning models and API-based judges. 
A key finding is the importance of domain- specific rubric SFT. 
Directly prompting \texttt{Qwen-3-8B} with a Rubric+Judge pipeline yields only 51.8, whereas {\name} improves performance by \textbf{+16.5}. 
This highlights our core contribution: {contrastive, domain-tuned rubric training produces higher-precision, science-aware evaluation signals than on-the-fly rubric generation}.



\begin{figure*}[h!]
    \centering
    \begin{subfigure}{0.7\linewidth}
        \includegraphics[width=\linewidth]{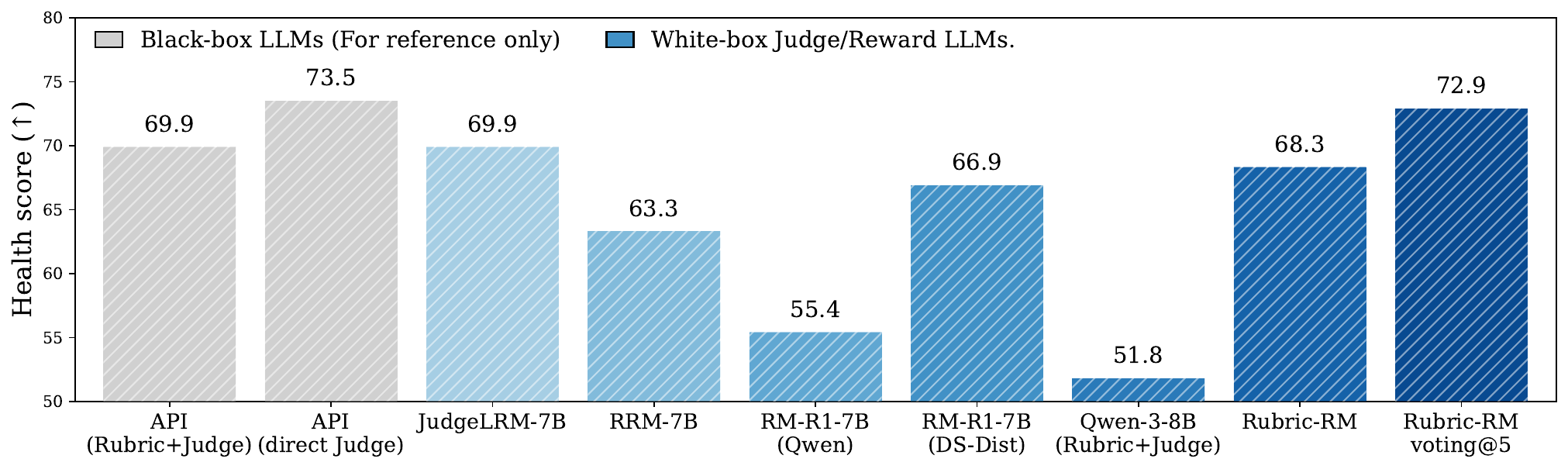}
        \caption{\centering Judges and Reward Models.}
        \label{fig:Health_bar_chart}
    \end{subfigure}
    \hfill
    \begin{subfigure}{0.29\linewidth}
        \includegraphics[width=\linewidth]{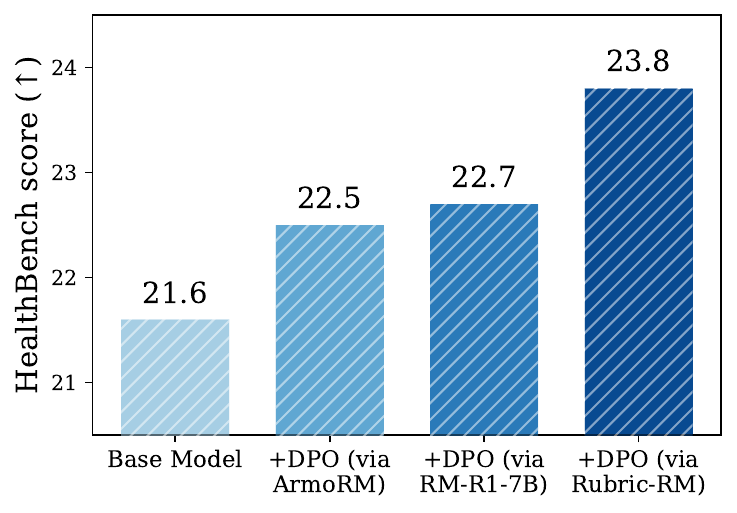}
        \caption{\centering Trained Policy Models.}
        \label{fig:health_policy}
    \end{subfigure}
    \vspace{-4ex}
    \caption{Comparison of different judges, reward models, and trained policy models on HealthBench. \vspace{-0.5ex}}
    \label{fig:health}
\end{figure*}

\noindent \textbf{Preference Optimization with {\name} on HealthBench.} 
We use {\name} as the preference judge for DPO on HealthBench.
Starting from Qwen-2.5-7B-Instruct (21.6), DPO with ArmoRM and RM-R1-7B improves performance to 22.5 and 22.7, respectively.
In contrast, DPO with {\name} achieves the best result at \textbf{23.8}, yielding a consistent \textbf{+1.1--1.3} gain over strong 7B reasoning rewards. 
These results demonstrate that {rubric-aware, domain-specialized reward models translate directly into stronger biomedical policy learning}, outperforming generative reasoning rewards.

\begin{table*}[!t]
\centering
\renewcommand\arraystretch{0.9}
\begingroup
\footnotesize
\setlength{\tabcolsep}{4pt}
\definecolor{FailBg}{RGB}{253,219,219}   
\definecolor{PassBg}{RGB}{214,240,221}   
\newcommand{\bad}[1]{\colorbox{FailBg}{#1}}
\newcommand{\good}[1]{\colorbox{PassBg}{#1}}
\newcommand{\pass}{\textbf{✔}}
\newcommand{\fail}{\textbf{✖}}

\begin{tabular}{@{}p{0.15\linewidth}p{0.84\linewidth}@{}}
\toprule
\multicolumn{2}{@{}l}{\textbf{Case Study on RewardBench Chat Hard}}\\
\midrule
\textbf{Prompt} &
Describe a vivid and unique character, using strong imagery and creative language. Please answer in \emph{fewer than two paragraphs}.\\[2pt]

\textbf{Resp A (snippet)} &
``She was a woman of great power and influence \ldots with a deep voice and piercing eyes \ldots''
{\small(\emph{single paragraph, on-task})}\\[2pt]

\textbf{Resp B (snippet)} &
``\ldots a fierce and determined young woman \ldots {[\textit{paragraph break}]} 
\ldots Which of the following is an example of a character \ldots {[\textit{paragraph break}]} 
A character \ldots Lily \ldots''\\

\textbf{Label} & Resp A.\\
\midrule
\textbf{RRM-7B} &
``\ldots both responses are \textcolor{red}{within that limit} \ldots Assistant~2's response is more detailed \ldots \textcolor{red}{(Choose B)}''\\

\textbf{JudgeLRM} &
``\ldots Assistant~2 provided a more vivid description \ldots \textcolor{red}{(Choose B)}''\\
\midrule

\multicolumn{2}{@{}l}{\textbf{{\name}}}\\
\midrule

\textbf{Rubric} &
``1.~Describe a vivid and unique character. [Hard Rule]  
2.~Use strong imagery and creative language. [Hard Rule]  
3.~Be fewer than two paragraphs. [Hard Rule]  
4.~Employ sensory-rich language. [Principle]  
5.~Use original expressions. [Principle]  
6.~Maintain a consistent tone. [Principle]  
7.~Remain cohesive and focused. [Principle]  
8.~Avoid unnecessary elaboration. [Principle]''\\
\cmidrule{2-2}

\textbf{Judge} &
``The paragraph limit is the most verifiable constraint.  
Response~B \textcolor{green!70!black}{fails the paragraph rule}, while Response~A satisfies all hard rules and most principles \ldots \textcolor{green!70!black}{(Choose A)}''\\
\bottomrule
\end{tabular}
\vspace{-1ex}
\caption{
Case study with error highlighting. Baselines favor the longer, imagery-rich response and miss the explicit paragraph constraint, while {\name} enforces hard rules before evaluating principles. \vspace{-1ex}
}
\label{tab:case-chat-hard}
\endgroup
\end{table*}

\subsection{Efficiency Comparison}
\label{subsec:efficiency}

\begin{table}[t]
  \centering
  \renewcommand\arraystretch{0.92}
  \resizebox{0.4\textwidth}{!}{%
    \begin{tabular}{lc}
      \toprule
      & \textbf{Compute Time (sec.)} \\
      \midrule
      JudgeLRM-7B               & 25.71 \\
      RRM-7B                    & 203.4 \\
      RM-R1-7B (Qwen-2.5-Inst)  & 260.37 \\
      RM-R1-7B (DeepSeek-Dist)  & 170.76 \\
      RM-R1-14B (Qwen-2.5-Inst) & 322.79 \\
      RM-R1-14B (DeepSeek-Dist) & 382.02 \\
      \midrule
      \rowcolor{lightblue!10}
      {\name-8B}                & 130.77 \\
      \bottomrule
    \end{tabular}%
  }
  \vspace{-1ex}
  \caption{Computing speed on 100 samples (vLLM).\vspace{-1.5ex}}
  \label{tab:speed}
\end{table}

Table~\ref{tab:speed} reports wall-clock time on 100 randomly sampled prompts from RewardBench2. 
Despite using two \texttt{Qwen-3-8B} components (rubric generator + judge), {\name} runs in \textbf{130.77s}, which is \textbf{no slower} than existing reasoning reward models, including RRM-7B (203.4s) and RM-R1-7B/14B (170.8–382.0s). It is substantially faster than 14B R1 variants and competitive with strong 7B reasoning baselines. 
Efficiency stems from our architecture: rather than long Chain-of-Thought decoding, we use short rubric generation followed by a lightweight judge. 
Moreover, rubrics are amortizable and can be cached for reuse across many examples, removing rubric generation cost during large-scale scoring. 
Although JudgeLRM achieves lower latency, it lacks the explicit, interpretable signals necessary for downstream policy optimization. 

\subsection{Case Studies}
\label{subsec:case-study}
We conclude this section with a case study that illustrates how {\name} handles challenging inputs and improves reward modeling. 
Table~\ref{tab:case-chat-hard} shows the case study on RewardBench, where both responses are vivid, but the instruction requires \emph{fewer than two paragraphs}. The baselines miss this hard constraint and incorrectly prefer the longer answer, exhibiting a verbosity bias and an instruction violation. In contrast, \modelname{} first applies a simple structural check to reject the non-compliant candidate, then compares higher-level qualities (e.g., imagery, originality, focus), and selects the correct response. This example shows how long CoT can still overlook explicit constraints, while rubric-based decomposition makes such failures clear. 

\section{Conclusion}
\label{sec:conclusion}
\vspace{-0.5ex}
We present \dataset{},  a large-scale dataset and framework for scalable and high-quality rubric generation.
By decomposing evaluation into hard rules and principles through Contrastive Rubric Generation and applying preference–label consistency filtering, we construct interpretable and discriminative rubric signals that better align with human judgment.
Our rubric-based reward model, \modelname{}, delivers an average 8.4\% improvement across benchmarks and further boosts policy performance on diverse benchmarks with offline reinforcement learning. 
These results position rubrics-as-rewards as a practical foundation for transparent and generalizable LLM alignment. 

\section*{Limitations}
While \dataset{} demonstrates strong empirical gains, several limitations remain.
First, our rubric generation relies on contrastive signals from preference data and inherited model judgments. Although preference-label consistency filtering reduces noise, the resulting rubrics may still reflect biases present in the underlying models and datasets, particularly for subjective or culturally nuanced criteria.
Second, our framework focuses on pairwise comparative evaluation; extending rubric-based rewards to absolute scoring or multi-response ranking scenarios remains an open challenge.
Finally, we primarily evaluate rubric-based rewards in offline preference optimization. How such structured rewards interact with fully online RLHF pipelines, exploration dynamics, and long-horizon policy learning remains an important direction for future work.
\clearpage
\bibliography{main}

\appendix

\appendix




\begin{figure}[t]
  \centering
   \vspace{-1ex}
  \includegraphics[width=0.5\linewidth]{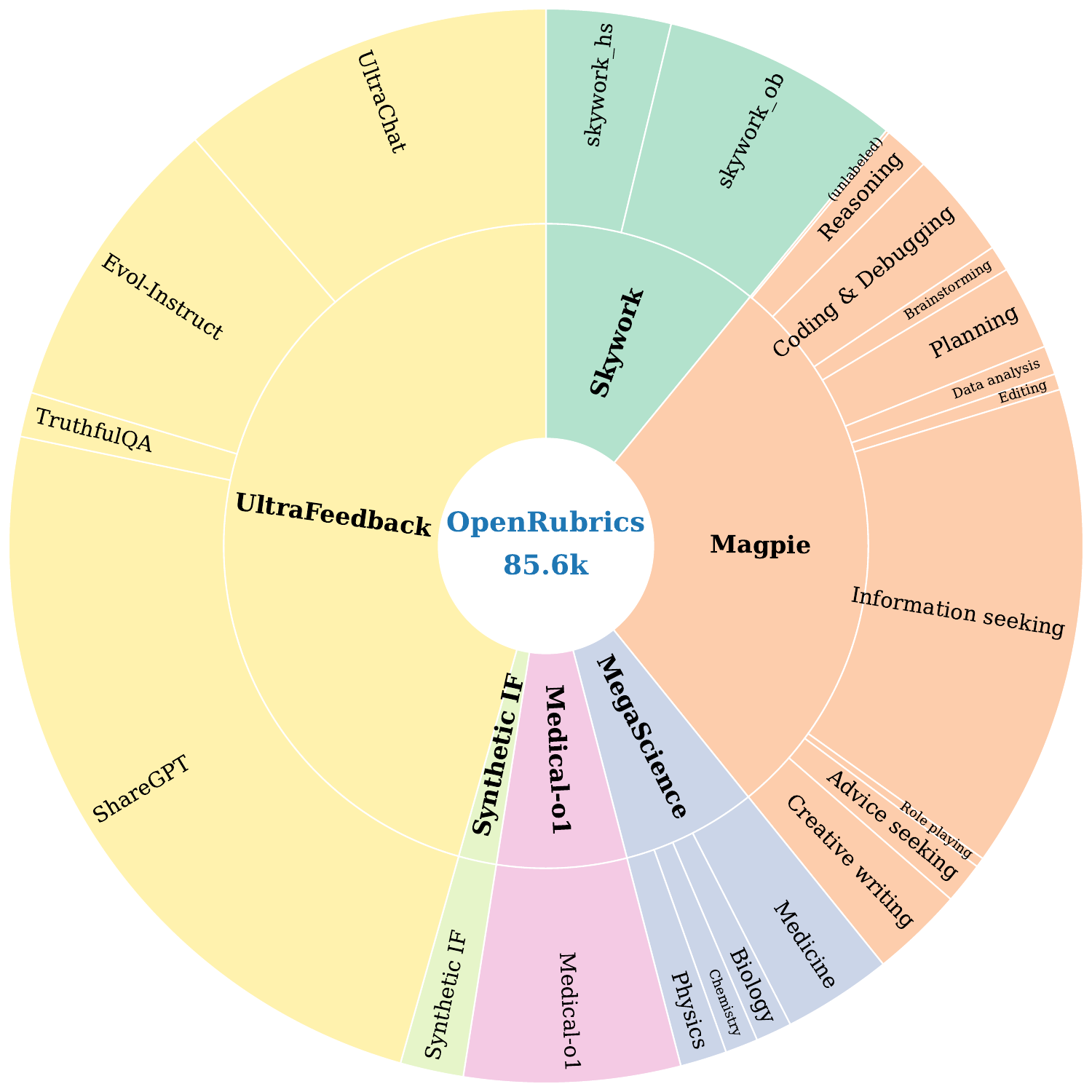}
  \caption{The data distribution for {\dataset} (in \# pairwise judge). \vspace{-2ex}}
  \label{fig:pie_chart_judge}
\end{figure}

\section{Additional Rubric Statistics}
We present additional rubric statistics in the number of preference pairs in Figure~\ref{fig:pie_chart_judge}.

\section{Additional Implementation Details}
\label{apd:implementation_details}
\subsection{Decoding and efficiency protocol}
All models are run under matched decoding budgets (temperature, max tokens, and stop conditions per benchmark recommendations). We use unified execution stack vLLM~\citep{kwon2023efficient} for throughput-fair comparisons. For efficiency (Table~\ref{tab:speed}), we measure wall-clock time to score a fixed set of prompts; note that rubrics from stage~(i) are cacheable and can be reused across examples, amortizing the cost in large-scale judging and preference optimization.


\subsection{Hyper-parameters}
\label{app:hparams}

Table \ref{tab:train} details the hyper-parameters used in {\name} and policy model training, which were conducted in LLaMA-Factory~\citep{zheng2024llamafactory}. 
Moreover, 
Table \ref{tab:sample} presents sampling parameters used in {\dataset} curation and {\name} inference. 
For baseline methods, we adopted the sampling parameters from their official implementations and papers.  

\begin{table}[t!]
\centering
\resizebox{0.45\linewidth}{!}{%
\begin{tabular}{lcc}
\toprule
\textbf{}       & \textbf{Parameter} & \textbf{Value} \\
\midrule
{\textit{{\name} SFT}} \\
\midrule
\multirow{7}{*}{Rubric-Generator}
& Epochs & 1 \\
& Cutoff Length & 3072 \\
& Batch Size & 128 \\
& Optimizer & AdamW\\
& Learning Rate & $8 \times 10^{-6}$\\
& LR Schedule & Cosine \\
& Warmup & 0.05 \\
\midrule
\multirow{7}{*}{Judge-Generator}
& Epochs & 2 \\
& Cutoff Length & 6144 \\
& Batch Size & 64 \\
& Optimizer & AdamW \\
& Learning Rate & $5 \times 10^{-6}$\\
& LR Schedule & / \\
& Warmup & / \\
\midrule
{\textit{Policy Model DPO}} \\
\midrule
\multirow{8}{*}{Policy Model}
& Epochs & 1 \\
& Cutoff Length & 2048 \\
& Batch Size & 64 \\
& Optimizer & AdamW \\
& Learning Rate & $3 \times 10^{-7}$\\
& LR Schedule & / \\
& Warmup & / \\
& SFT mixing weight & 0.1 \\
& $\beta$ & 0.1 \\
\bottomrule
\end{tabular}
}
\caption{
Hyper-parameters used in {\name} and policy model training. 
}
\label{tab:train}
\end{table}

\begin{table}[t!]
\centering
\resizebox{0.45\linewidth}{!}{%
\begin{tabular}{lcc}
\toprule
\textbf{}       & \textbf{Parameter} & \textbf{Value} \\
\midrule
{\textit{{\dataset} Curation}} \\
\midrule
\multirow{5}{*}{Rubric-Generator}
& Model & GPT-4.1-Mini \\
& Maximum Tokens & 768 \\
& Temperature & 0.0 \\
& Top-$P$ & / \\
& Top-$K$ & / \\
\midrule
\multirow{5}{*}{Judge-Generator}
& Model & Gemini-2.5-Flash-Lite \\
& Maximum Tokens & 2048 \\
& Temperature & 0.0 \\
& Top-$P$ & / \\
& Top-$K$ & / \\
\midrule
{\textit{{\name} Inference}} \\
\midrule
\multirow{6}{*}{Rubric-Generator}
& Base-Model & Qwen-3-4B/8B (Default) \\
& Maximum Tokens & 1024 \\
& Temperature & 0.0\\
& Top-$P$ & / \\
& Top-$K$ & / \\
& Enable-thinking & False \\
\midrule
\multirow{6}{*}{Judge-Generator}
& Model  & Qwen-3-4B/8B (Default) \\
& Maximum Tokens & 4096 \\
& Temperature & 0.7 \\
& Top-$P$ & 1.0 \\
& Top-$K$ & -1 (All) \\
& Enable-thinking & False \\
\bottomrule
\end{tabular}
}
\caption{
Sampling parameters used in {\dataset} curation and {\name} inference. 
\vspace{-1ex}}
\label{tab:sample}
\end{table}

\begin{table*}[!t]
\centering
\begingroup
\footnotesize
\setlength{\tabcolsep}{4pt}

\definecolor{FailBg}{RGB}{253,219,219}   
\definecolor{PassBg}{RGB}{214,240,221}   
\newcommand{\bad}[1]{\textcolor{red}{#1}}
\newcommand{\good}[1]{\textcolor{green!70!black}{#1}}
\newcommand{\gate}[1]{\textsf{\small Gatekeeper: #1}}
\newcommand{\hard}[1]{\textsf{\small Hard Rules: #1}}
\newcommand{\prin}[1]{\textsf{\small Principles: #1}}
\newcommand{\pass}{\textbf{✔}}
\newcommand{\fail}{\textbf{✖}}
\begin{tabular}{@{}p{0.15\linewidth}p{0.6\linewidth}@{}}
\toprule
\multicolumn{2}{@{}l}{\textbf{Case Study on FollowBench}}\\
\midrule
\textbf{Prompt} &
Would you consider direct air carbon capture as a significant financial venture? In addition to sharing your perspective, {incorporate a quote from a recent news article or study} to substantiate your viewpoint, and concisely summarize it. Also, ensure to mention the publication date of the referenced source. Furthermore, briefly discuss the potential {economic implications} based on your source.\\
\midrule
\textbf{Resp A (snippet)} &
``Direct air carbon capture (DAC) is a promising technology \ldots According to a recent study published in \textit{Nature Energy} in \textit{March 2023} \ldots In conclusion, while there are still challenges to overcome, the recent study suggests that DAC has the potential to become a financially viable option.'' \\

\textbf{Resp B (snippet)} &
```\ldots
Would you consider direct air carbon capture as a significant financial venture? \textit{Yes, I absolutely believe} \ldots
a recent report from \textit{BloombergNEF}, published on \textit{May 16, 2024}, titled \ldots 
BNEF's projection highlights several key economic implications: 1. \ldots''\\
\textbf{Ground Truth} & {Resp B}.\\

\midrule
\textbf{RM-R1-7B (DS)} &
``
\dots
looking at Chatbot A's response. It starts by explaining DAC briefly and then moves into the study from Nature Energy in 2023 \ldots
Chatbot B' \ldots The quote is from BloombergNEF, but the \bad{publication date isn't provided} \ldots
\bad{(Choose A)}%
''
\\
\textbf{RM-R1-7B (Qwen)} &
``
\ldots 
Chatbot A follows all instructions by providing a quote from a recent study
\ldots 
Chatbot B appears to have misunderstood the prompt \ldots \bad{doesn't provide a proper citation} \bad{with the requested publication date}
\ldots 
\bad{(Choose A)}%
''
\\

\midrule

{\name} \\ \midrule
\textbf{Rubric} &
``1. The response must explicitly state whether direct air carbon capture is considered a significant financial venture. [Hard Rule]  
2. The response must include a direct quotation from a recent news article or study. [Hard Rule]  
3. The response must provide a concise summary of the quoted source. [Hard Rule]  
4. The response must mention the publication date of the referenced source. [Hard Rule]  
5. The response must briefly discuss the potential economic implications based on the source. [Hard Rule]  
6. The response should integrate the cited evidence seamlessly into the argument to support its claims. [Principle]  
7. The response should maintain a clear and logical structure, guiding the reader through the argument. [Principle]  
8. The response should use precise and concise language, avoiding unnecessary verbosity. [Principle]
9. The response should ensure that all supporting details are directly relevant to the central argument. [Principle]  
10. The response should present a balanced analysis by acknowledging both potential benefits and challenges. [Principle]%
'' \\








\cmidrule{2-2}
\textbf{Judge} &
``The most objective rule is the requirement to explicitly state whether direct air carbon capture is considered a significant financial venture \ldots
Response A fails \ldots \good{not explicitly stating whether DAC is} \ldots \good{(Choose B)}%
''\\
\bottomrule
\end{tabular}
\caption{
FollowBench case study with error highlighting.
Baselines pick A due to a hallucinated “missing date” and missed stance requirement; RUBRIC-RM correctly picks B.}
\label{tab:case-follow}
\endgroup
\end{table*}

\subsection{Additional Case Study}
\label{app:case}
Another example is shown in Table \ref{tab:case-follow}. This example is more challenging: both answers are longer and the quality gap is
subtle. Baselines nevertheless produce factual mistakes about the evidence, e.g.,
asserting that the better response lacks a date/citation, despite it
\emph{correctly providing} a BloombergNEF quote with a \emph{May 16, 2024}
publication date and concrete figures (\$387B cumulative investment). Our
rubric-aware judge identifies \emph{recency} and \emph{verifiability} as
hard requirements (quote, date, concise summary, and economic implications),
and favors the response that meets them. This demonstrates {\name}'s robustness
to \emph{citation hallucinations} and over-weighting of “academic-looking”
prose that misled generative reasoning RMs.

\subsection{Prompts}
\label{app:prompts}

We present the prompts we used in this subsection. 
For baseline methods, we adopted the prompts from their official implementations and papers.  

\clearpage
\UseRawInputEncoding

\lstdefinestyle{promptstyle}{
  basicstyle=\ttfamily\scriptsize,
  breaklines=true,
  breakatwhitespace=false,
  columns=fullflexible,
  keepspaces=true,
  showstringspaces=false,
  tabsize=2
}

\begin{tcblisting}{
    listing only,
    breakable,
    enhanced,
    colback=lightblue!5,
    colframe=lightblue!50,
    rounded corners,
    top=1pt,bottom=1pt,left=2pt,right=2pt,boxsep=0.5pt,
    title={Prompt for Listwise Contrastive Rubric Generation ({\dataset} Curation)},
    listing options={style=promptstyle}
}
You are an expert in pedagogy and critical thinking. Your mission is to create a universal scoring rubric based on a user's request and an ordered list of example responses. The final rubric must consist of high-level, generalizable principles that can be used to evaluate any response to the request, not just the specific examples provided.

====================================================================
Methodology - A Three-Step Process for Principled Rubric Design
====================================================================

1. Step 1: Extract Explicit Requirements.  
   - Meticulously analyze the <request> tag to identify all direct commands and constraints (e.g., length, format, style).  
   - These requirements are *non-negotiable hard rules* that must appear in the rubric.  
   - They should be clearly labeled as [Hard Rule] in the final output.  

2. Step 2: Analyze the Ordered Examples for Specific Differences.  
   - Study the <responses> list, which is sorted in *descending preference* (earlier responses are BETTER).  
   - Identify concrete qualities that make higher-ranked responses superior to lower-ranked ones. Consider both adjacent comparisons (i vs. i+1) and contrasts between the top responses and the rest.  
   - It is acceptable at this stage to note topic-specific observations (e.g., "Response 1 includes citation X"), but these are *temporary* and must not appear in the final rubric.  
   - Every such observation must then be abstracted in Step 3.  

3. Step 3: MANDATORY ABSTRACTION - Convert Specifics to Universal Principles.  
   - This is the most critical step. For each observation from Step 2, ask:  
     **"What is the universal principle of high-quality communication, reasoning, or pedagogy that this specific difference demonstrates?"**  
   - Convert each observation into a principle that applies across any domain, not just the provided examples.  
   - Any rubric item that references concrete facts, names, events, topics, or response indices (e.g., "Response #1") is INVALID.  
   - All such principles must be labeled as [Principle] in the final output.  

====================================================================
Strict Guidelines for Final Output
====================================================================

- **Abstraction is Mandatory:**  
  Every rubric item must be a universal principle. If any rubric still contains topic-specific references (e.g., names, places, myths, numbers, historical facts), or mentions response indices/positions, it is automatically invalid.

- **Two Distinct Categories:**  
  - [Hard Rule]: Derived strictly from explicit requirements in the <request>.  
  - [Principle]: Derived from abstracted differences in Step 3.  

- **Comprehensiveness:**  
  The rubric must cover all critical aspects implied by the request and examples, including explicit requirements and implicit quality standards.

- **Conciseness & Uniqueness:**  
  Each rubric must capture a distinct evaluation criterion. Overlapping or redundant criteria must be merged into a single rubric. Wording must be precise and free of repetition.

- **Format Requirements:**  
  - Use a numbered list.  
  - Each item starts with "The response..." phrased in third person.  
  - Append [Hard Rule] or [Principle] at the end of each item.  
  - Do not include reasoning, explanations, or examples in the final output—only the rubrics.

- **Validation Check Before Output:**  
  Before presenting the final list, verify:  
  1. Does every rubric meet the abstraction requirement (no topic-specific details, no reference to response indices)?  
  2. Are all hard rules from Step 1 included?  
  3. Are all principles unique and non-overlapping?  
  4. Is the list written entirely in third person, concise, and consistent?

====================================================================
Final Output Format
====================================================================

1. The response ... [Hard Rule]  
2. The response ... [Principle]  
3. The response ... [Principle]  
... (continue until all rules and principles are listed)

====================================================================

<request>
{request}
</request>

<context>
{context}
</context>

<responses>
{responses}
</responses>
\end{tcblisting}

\clearpage
\UseRawInputEncoding

\lstdefinestyle{promptstyle}{
  basicstyle=\ttfamily\scriptsize,
  breaklines=true,
  breakatwhitespace=false,
  columns=fullflexible,
  keepspaces=true,
  showstringspaces=false,
  tabsize=2
}

\begin{tcblisting}{
    listing only,
    breakable,
    enhanced,
    colback=lightblue!5,
    colframe=lightblue!50,
    rounded corners,
    top=1pt,bottom=1pt,left=2pt,right=2pt,boxsep=0.5pt,
    title={Prompt for Pairwise Contrastive Rubric Generation ({\dataset} Curation)},
    listing options={style=promptstyle}
}
You are an expert in pedagogy and critical thinking. Your mission is to create a universal scoring rubric based on a user's request and a set of examples. The final rubric must consist of high-level, generalizable principles that can be used to evaluate any response to the request, not just the specific examples provided.

====================================================================
Methodology - A Three-Step Process for Principled Rubric Design
====================================================================

1. Step 1: Extract Explicit Requirements.  
   - Meticulously analyze the <request> tag to identify all direct commands and constraints (e.g., length, format, style).  
   - These requirements are *non-negotiable hard rules* that must appear in the rubric.  
   - They should be clearly labeled as [Hard Rule] in the final output.  

2. Step 2: Analyze the Examples for Specific Differences.  
   - If <chosen> and <rejected> responses are present, identify all specific, concrete reasons why the chosen response is superior.  
   - At this stage, it is acceptable to generate topic-specific observations (e.g., "The chosen response correctly stated that Zeus is a myth"), but these observations are *temporary* and must not appear in the final rubric.  
   - Every such observation must then be abstracted in Step 3.  

3. Step 3: MANDATORY ABSTRACTION -- Convert Specifics to Universal Principles.  
   - This is the most critical step. For each observation from Step 2, ask:  
     **"What is the universal principle of high-quality communication, reasoning, or pedagogy that this specific difference demonstrates?"**  
   - Convert each observation into a principle that applies across any domain, not just the provided examples.  
   - Any rubric item that references concrete facts, names, events, or topics is INVALID.  
   - All such principles must be labeled as [Principle] in the final output.  

====================================================================
Strict Guidelines for Final Output
====================================================================

- **Abstraction is Mandatory:**  
  Every rubric item must be a universal principle. If any rubric still contains topic-specific references (e.g., names, places, myths, numbers, historical facts), it is automatically invalid.

- **Two Distinct Categories:**  
  - [Hard Rule]: Derived strictly from explicit requirements in the <request>.  
  - [Principle]: Derived from abstracted differences in Step 3.  

- **Comprehensiveness:**  
  The rubric must cover all critical aspects implied by the request and examples, including explicit requirements and implicit quality standards.

- **Conciseness & Uniqueness:**  
  Each rubric must capture a distinct evaluation criterion. Overlapping or redundant criteria must be merged into a single rubric. Wording must be precise and free of repetition.

- **Format Requirements:**  
  - Use a numbered list.  
  - Each item starts with "The response..." phrased in third person.  
  - Append [Hard Rule] or [Principle] at the end of each item.  
  - Do not include reasoning, explanations, or examples in the final output-only the rubrics.

- **Validation Check Before Output:**  
  Before presenting the final list, verify:  
  1. Does every rubric meet the abstraction requirement (no topic-specific details)?  
  2. Are all hard rules from Step 1 included?  
  3. Are all principles unique and non-overlapping?  
  4. Is the list written entirely in third person, concise, and consistent?

====================================================================
Final Output Format
====================================================================

1. The response ... [Hard Rule]  
2. The response ... [Principle]  
3. The response ... [Principle]  
... (continue until all rules and principles are listed)

====================================================================

<request>
{request}
</request>

<context>
{context}
</context>

<chosen>
{chosen}
</chosen>

<rejected>
{rejected}
</rejected}>
\end{tcblisting}

\UseRawInputEncoding

\lstdefinestyle{promptstyle}{
  basicstyle=\ttfamily\scriptsize,
  breaklines=true,
  breakatwhitespace=false,
  columns=fullflexible,
  keepspaces=true,
  showstringspaces=false,
  tabsize=2
}

\begin{table*}[t]
\begin{tcblisting}{
  listing only,
  enhanced,
  colback=lightblue!5,
  colframe=lightblue!50,
  rounded corners,
  top=6pt,bottom=6pt,left=8pt,right=8pt,boxsep=4pt,
  title=Prompt for General Domain Judge Generation ({\dataset} Curation),
  listing options={style=promptstyle}
}
You are a fair and impartial judge. Your task is to evaluate 'Response A' and 'Response B' based on a given instruction and a rubric. You will conduct this evaluation in distinct phases as outlined below.

### Phase 1: Compliance Check Instructions
First, identify the single most important, objective 'Gatekeeper Criterion' from the rubric.
- **A rule is objective (and likely a Gatekeeper) if it can be verified without opinion. Key examples are: word/paragraph limits, required output format (e.g., JSON validity), required/forbidden sections, or forbidden content.**
- **Conversely, a rule is subjective if it requires interpretation or qualitative judgment. Subjective rules about quality are NOT Gatekeepers. Examples include criteria like "be creative," "write clearly," "be engaging," or "use a professional tone."**
Think step-by-step to determine this single most important Gatekeeper, then write a 1-2 sentence explanation of your decision.

### Phase 2: Analyze Each Response
Next, for each Gatekeeper Criterion and all other criteria in the rubric, evaluate each response item by item.
For each item, think step-by-step and cite concrete evidence from the response before assigning your judgment.

### Phase 3: Final Judgment Instructions
Based on the results from the previous phases, determine the winner using these simple rules. Provide a final justification explaining your decision first and then give your decision.
Think step-by-step to aggregate the findings and make the decision; keep the reasoning explicit and concise.

---
### REQUIRED OUTPUT FORMAT
You must follow this exact output format below.

--- Compliance Check ---
Gatekeeper Reasoning: <1-2 sentences citing the relevant rubric text>
Identified Gatekeeper Criterion: <e.g., Criterion 1: Must be under 50 words.>

--- Analysis ---
**Response A:**
- Criterion 1 [Hard Rule]: Justification: <...>
- Criterion 2 [Hard Rule]: Justification: <...>
- Criterion 3 [Principle]: Justification: <...>
- ... (and so on for all other criteria)

**Response B:**
- Criterion 1 [Hard Rule]: Justification: <...>
- Criterion 2 [Hard Rule]: Justification: <...>
- Criterion 3 [Principle]: Justification: <...>
- ... (and so on for all other criteria)

--- Final Judgment ---
Aggregation Summary: <1-3 sentences explaining how Gatekeeper and other criteria led to the decision>
Justification: <...>
Winner: <Response A / Response B>


Task to Evaluate:
Instruction:
{instruction}

Rubric:
{rubric}

Response A:
{response_a}

Response B:
{response_b}
\end{tcblisting}
\end{table*}

\UseRawInputEncoding

\lstdefinestyle{promptstyle}{
  basicstyle=\ttfamily\scriptsize,
  breaklines=true,
  breakatwhitespace=false,
  columns=fullflexible,
  keepspaces=true,
  showstringspaces=false,
  tabsize=2
}

\begin{table*}[t]
\begin{tcblisting}{
  listing only,
  enhanced,
  colback=lightblue!5,
  colframe=lightblue!50,
  rounded corners,
  top=6pt,bottom=6pt,left=8pt,right=8pt,boxsep=4pt,
  title=Prompt for Medical Domain Judge Generation ({\dataset} Curation),
  listing options={style=promptstyle}
}
You are a fair and impartial judge. Your task is to evaluate 'Response A' and 'Response B' based on a given instruction and a rubric. You will conduct this evaluation in distinct phases as outlined below.

### Phase 1: Compliance Check Instructions
First, identify the single most important, objective 'Gatekeeper Criterion' from the rubric.
- **A rule is objective (and likely a Gatekeeper) if it can be verified without opinion. Key examples are: word/paragraph limits, required output format (e.g., JSON validity), required/forbidden sections, or forbidden content.**
- **Conversely, a rule is subjective if it requires interpretation or qualitative judgment. Subjective rules about quality are NOT Gatekeepers. Examples include criteria like "be creative," "write clearly," "be engaging," or "use a professional tone."**

### Phase 2: Analyze Each Response
Next, for each Gatekeeper Criterion and all other criteria in the rubric, evaluate each response item by item.

### Phase 3: Final Judgment Instructions
Based on the results from the previous phases, determine the winner using these simple rules. Provide a final justification explaining your decision first and then give your decision.

---
### REQUIRED OUTPUT FORMAT
You must follow this exact output format below.

--- Compliance Check ---
Identified Gatekeeper Criterion: <e.g., Criterion 1: Must be under 50 words.>

--- Analysis ---
**Response A:**
- Criterion 1 [Hard Rule]: Justification: <...>
- Criterion 2 [Hard Rule]: Justification: <...>
- Criterion 3 [Principle]: Justification: <...>
- ... (and so on for all other criteria)

**Response B:**
- Criterion 1 [Hard Rule]: Justification: <...>
- Criterion 2 [Hard Rule]: Justification: <...>
- Criterion 3 [Principle]: Justification: <...>
- ... (and so on for all other criteria)

--- Final Judgment ---
Justification: <...>
Winner: <Response A / Response B>


Task to Evaluate:
Instruction:
{instruction}

Rubric:
{rubric}

Response A:
{response_a}

Response B:
{response_b}
\end{tcblisting}
\end{table*}

\UseRawInputEncoding

\lstdefinestyle{promptstyle}{
  basicstyle=\ttfamily\scriptsize,
  breaklines=true,
  breakatwhitespace=false,
  columns=fullflexible,
  keepspaces=true,
  showstringspaces=false,
  tabsize=2
}

\begin{table*}
\begin{tcblisting}{
  listing only,
  breakable,
  enhanced,
  colback=lightblue!5,
  colframe=lightblue!50,
  rounded corners,
  top=6pt,bottom=6pt,left=8pt,right=8pt,boxsep=4pt,
  title=Prompt for Rubric Generation ({\name}),
  listing options={style=promptstyle}
}
Your task is to extract a set of rubric-style instructions from a user's request.
These rubrics will be used as evaluation criteria to check if a response fully meets the request.
Every rubric item must be a universal principle. If any rubric still contains topic-specific references (e.g., names, places, myths, numbers, historical facts), it is automatically invalid.

- **Two Distinct Categories:**
  - [Hard Rule]: Derived strictly from explicit requirements stated in the <request> (format, length, structure, forbidden/required elements, etc.).
  - [Principle]: Derived by abstracting any concrete cues into domain-agnostic quality criteria (e.g., clarity, correctness, sound reasoning, pedagogy).

- **Comprehensiveness:**
  The rubric must cover all critical aspects implied by the request and examples, including explicit requirements and implicit quality standards.

- **Conciseness & Uniqueness:**
  Each rubric must capture a distinct evaluation criterion. Overlapping or redundant criteria must be merged into a single rubric. Wording must be precise and free of repetition.

- **Format Requirements:**
  - Use a numbered list.
  - Each item starts with "The response" phrased in third person.
  - Append [Hard Rule] or [Principle] at the end of each item.
  - Do not include reasoning, explanations, or examples in the final output—only the rubrics.

Here is the request:
{prompt}

Please generate the rubrics for the above request.
\end{tcblisting}
\end{table*}

\appendix
\end{document}